\documentclass[11pt,a4paper]{article}
\usepackage[hyperref]{eacl2021}
\usepackage{times}
\usepackage{latexsym}
\usepackage{algorithm}
\usepackage{algpseudocode}
\usepackage{todonotes}

\usepackage{multirow}
\usepackage{subcaption}
\usepackage{booktabs}

\usepackage{soul}
\newcommand{\indomain}[1]{$\mathcal{D}^{#1}_{i}$}
\newcommand{\outofdomain}[1]{$\mathcal{D}^{#1}_{o}$}
\newcommand{\traindata}[1]{$\mathcal{D}^{#1}_{train}$}
\newcommand{\model}[1]{$\mathcal{M}^{#1}$}
\newcommand{\schedule}{$\mathcal{S}$}

\newcommand{\best}[1]{\textbf{#1}}

\usepackage{microtype}

\aclfinalcopy % Uncomment this line for the final submission
\title{Gradual Fine-Tuning for Low-Resource Domain Adaptation}

\newcommand\Mark[1]{\textsuperscript#1}

\author{Haoran Xu\Mark{1}, Seth Ebner\Mark{1}, 
Mahsa Yarmohammadi\Mark{1}, Aaron Steven White\Mark{2},\\ 
{\bf Benjamin Van Durme\Mark{1}}, 
{\bf Kenton Murray\Mark{1}}\\[1em]
\Mark{1}Johns Hopkins University, \Mark{2}University of Rochester\\[1em]
\texttt{\{hxu64,seth,mahsa,vandurme,kenton\}@jhu.edu,}\\
\texttt{aaron.white@rochester.edu}\\[1em]
}
\date{}

\begin{document}
\maketitle
\begin{abstract}
%Fine-tuning is known to improve NLP models, adapting to a target domain where there is less data from an initial model trained on more plentiful but less domain-salient examples.
Fine-tuning is known to improve NLP models by adapting an initial model trained on more plentiful but less domain-salient examples to data in a target domain.
%Fine-tuning NLP models improves performance on a target domain by learning from a related domain with more data to provide good initialization that helps the model fit the target domain.
Such domain adaptation is typically done using one stage of fine-tuning.  We demonstrate that gradually fine-tuning in a multi-stage process can yield substantial further gains and can be applied without modifying the model or learning objective.
% Fine-tuning pre-trained NLP models has been shown to increase performance on specific tasks different from the original data and model. Here we show that a one-step fine-tuning approach is not the optimal method on Information Extraction tasks. Rather, we show that gradually fine-tuning in a multi-step process, similar to curriculum learning, can yield substantial gains.
\end{abstract}

\section{Introduction}
Domain adaptation is a technique for practical applications in which one wants to learn a model for a task in a particular domain with too few instances of in-domain data to directly learn a model. Common approaches for domain adaptation make use of fine-tuning \citep{dabre2019exploiting,li-specia-2019-improving,imankulova-etal-2019-exploiting}, in which a model is pretrained on a large amount of out-of-domain but task-relevant data and then refined toward the target domain by subsequently training on in-domain data. This fine-tuning procedure is often performed in one stage: the pretrained model is trained on the in-domain data until convergence~\citep{chu-etal-2017-empirical,min-etal-2017-question}. We propose a \textit{gradual} fine-tuning approach, in which a model is iteratively trained to convergence on data whose distribution progressively approaches that of the in-domain data. Intuitively, the model is eased toward the target domain rather than abruptly shifting to it.

%Text augmentation \cite{zhang2015character,coulombe2018text} has been playing an imperative role in NLP tasks,  especially when clean and high-quality labeled data is inadequate to train a satisfactory learner for specific tasks. Recently, increasing number of researches prefer to look for enormous size of out-of-domain data to build a pre-trained model trained by large dataset for specific applications and fine-tune the model on in-domain data. This is to ensure the distribution of the model prediction would be consistent with the target domain, e.g., claim detection \citep{chakrabarty-etal-2019-imho} and NMT \citep{li-specia-2019-improving}. Thus, the model can transfer from the general domain to the target domain, and benefit from the larger training dataset they have seen to perform better.

\begin{figure}[t]
    \centering
    \includegraphics[width=7.5cm]{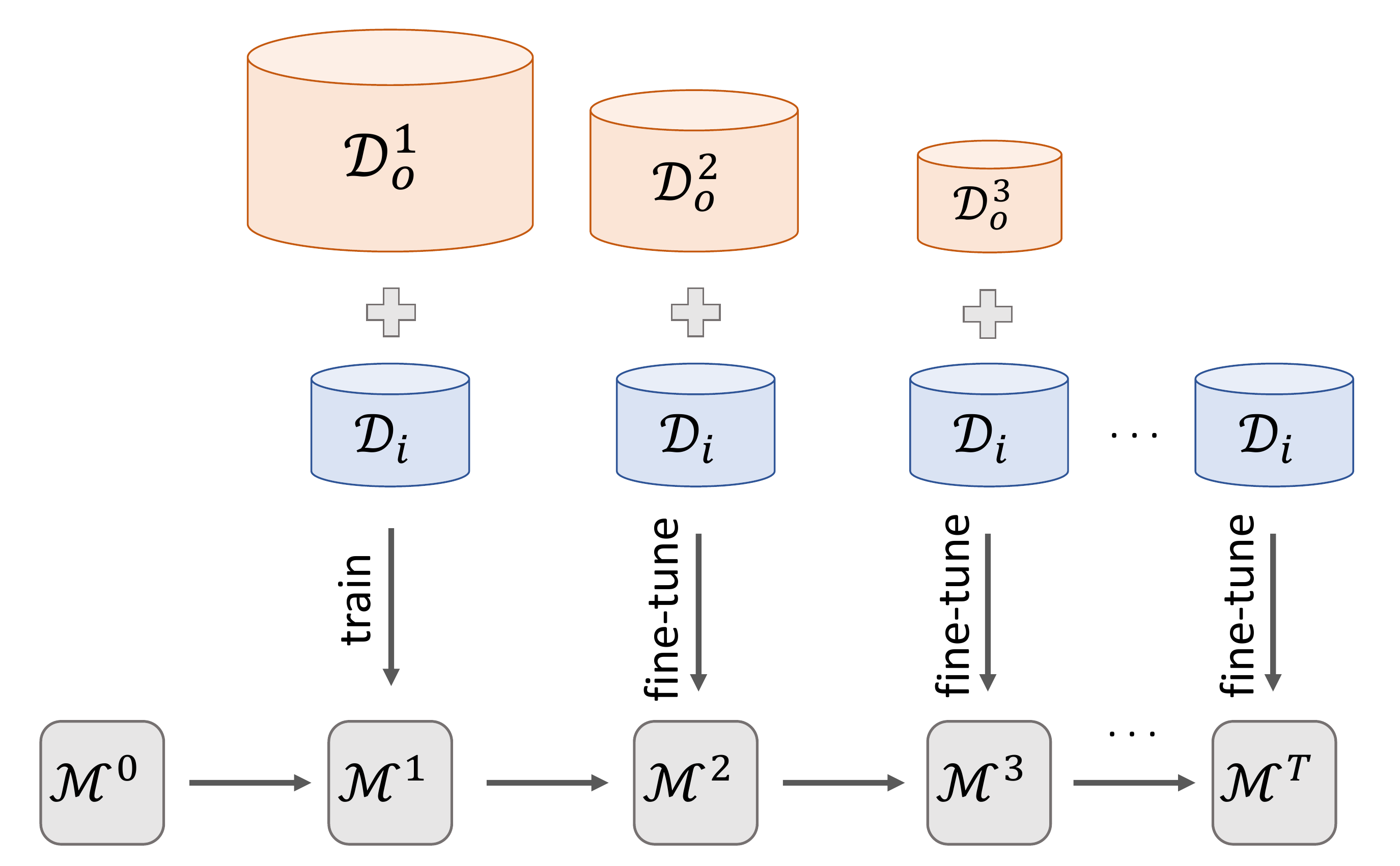}
    \caption{Stages of gradual fine-tuning: 1) Train the model, \model{}, on a mixture of in-domain data, \indomain{}, and out-of-domain data, \outofdomain{}; 2) iteratively fine-tune on mixed domain data with decreasing amounts of out-of-domain data; 3) fine-tune on only in-domain data.} %bvd Superscripts indicate time steps.}
    \label{fig:gradual}
\end{figure}

% \st{Domain transfer is usually conducted in one-step approach --- directly fine-tuning the pretrained model on the in-domain data. We show that the performance of multi-step gradual fine-tuning surpasses that of one-step fine-tuning on downstream tasks.}
Inspired by the general approach of curriculum learning of training a model on a trajectory from easier instances to more difficult instances~\citep{10.1145/1553374.1553380}, we train a model on a sequence of datasets, each of which would be increasingly difficult to learn on its own due to its size. Each dataset in the sequence interpolates between the data in the previous iteration and the target domain data. We hypothesize that just as in curriculum learning where first learning from easier instances helps models subsequently learn from more difficult instances, the interpolation process yields datasets with distributions that are increasingly useful for helping models learn during subsequent stages of the gradual fine-tuning procedure. We begin by training the model on data that contains a mix of out-of-domain and in-domain instances, then increase the concentration of in-domain data in each fine-tuning stage. In this way, at each stage of fine-tuning we increase the similarity between the current domain and the target domain, which enables the model to potentially better fit the distribution of the target domain. The approach is illustrated in \autoref{fig:gradual}. %The gradual fine-tuning process is depicted in.

We conduct experiments on two NLP tasks to demonstrate the effectiveness of gradual fine-tuning. We first look at (1) dialogue state tracking with the MultiWOZ v2.0 dataset \citep{budzianowski2018large}, which is a collection of human-to-human conversation transcriptions in multiple domains. We focus on single dialogue domains and utilize out-of-domain data to improve accuracy of slot classification in the target domain. %bvd via gradual fine-tuning.
We then consider (2) an event extraction task from the ACE 2005 dataset (LDC2006T06) for which we augment the Arabic target domain data with English data.

Gradual fine-tuning is also simple to implement. If fine-tuning is already supported by the code, then only new configuration files need to be created to specify the (amount of) data used at each iteration.\footnote{One may also need to implement data subsampling.} No adjustments to model or training code are needed. Just by modifying the training approach, one can obtain substantial improvements.\footnote{Supporting code: \url{https://github.com/fe1ixxu/Gradual-Finetune}.}

%We then consider an event extraction task from the IARPA BETTER program\footnote{\url{https://www.iarpa.gov/index.php/research-programs/better}} with no off-the-shelf out-of-domain data. After describing how to recast a source of out-of-domain data, we show that gradual fine-tuning substantially boosts task performance.

% The first experiment is specific, where we consider a complex Abstract Event extraction task with no off-the-shelf out-of- domain data,  and we give an example of how to find out-of-domain data and make it be compatible with the single task model. The second experiment develops on Multi-Domain Wizard-of-Oz dataset (MultiWOZ) \citep{budzianowski2018large}, which contains human-human written and fully labeled dialogue dataset in multiply domains. This task is more general, and the out-of-domain dataset is freely available for one of target domains. Architectures of models for the both above experiments are based on the state-of-the-art Universal Decompositional Semantics (UDS) parser \citep{stengel2020universal}.

\section{Related Work}
%For a particular task in a target domain, we may possess insufficient labeled data but have a comparatively large amount of labeled data in a more general domain. 

%For a target learning task $T_T$, we may only possess limited labeled data in the target domain $D_T$ but have a large amount of labeled data in a separate source domain $D_S$ ($D_S \neq D_T$). Transfer fine-tuning aims to improve the target learning task by leveraging the information learned from $D_S$ and $T_S$.

% Prior work has shown that performance on downstream tasks can be greatly improved through \textit{inductive transfer} \citep{pan2009survey,Ruder2019Neural} compared with training from scratch. Inductive transfer involves pretraining on a task other than the downstream task, such as fine-tuning pretrained language models on specific tasks \citep{howard-ruder-2018-universal,cui-etal-2019-fine,arase2019transfer,DBLP:conf/icml/HoulsbyGJMLGAG19,zhang2020can}.

\citet{howard-ruder-2018-universal} propose an effective inductive transfer learning method for language model fine-tuning and demonstrate improvements on text classification tasks. \citet{gururangan-etal-2020-dont} also show improvements on target task performance by fine-tuning pretrained language models on in-domain data and on the target task's training data. In this work, we focus on adapting the entire model, not just the underlying language model encoder. Our approach is a form of transductive transfer \citep{pan2009survey}, in which the pretraining and fine-tuning tasks are the same. In the transductive transfer setting, we hope to learn task-specific information by training on a large (potentially out-of-domain) dataset, and then subsequently adjust model parameters based on domain-specific information learned from in-domain data.

Transductive transfer has been effective for tasks such as question answering \citep{min2017question}, machine translation \citep{sennrich2015improving}, and open information extraction \citep{sarhan2020can}. \citet{wu-etal-2019-transferable} fine-tune toward a target domain for dialogue state tracking using Gradient Episodic Memory~\cite{lopez2017gradient} to avoid catastrophic forgetting~\cite{mccloskey1989catastrophic}. \citet{NEURIPS2019_2c3ddf4b} introduce an uncertainty-based regularization method to overcome catastrophic
forgetting in the continual learning setting. There have also been successful approaches for cross-lingual information extraction and semantic role labeling using no target language data~\cite{subburathinam-etal-2019-cross}, mixed source and translated target language data~\cite{fei2020cross}, and language-independent model transfer~\cite{daza-frank-2019-translate,9165903}.

% Recently, \citet{sarhan2020can} study in domain adaptation to improve the performance of open information extraction in a bio-medical domain by transferring from a high-resource news domain.

%\citet{daume2009frustratingly} propose a feature augmentation approach to domain adaptation, which can be extended to semi-supervised learning~\cite{daume-iii-etal-2010-frustratingly} and to models with dense representations~\cite{kim-etal-2016-frustratingly}.

\citet{jiang-zhai-2007-instance} propose a method for upweighting the importance of target domain instances relative to source domain instances to improve domain adaptation. The iterative increase in concentration of target domain data in the mixed domain data used in gradual fine-tuning can be seen as analogous to giving target domain instances more weight.
%Adversarial approaches have also been effective for domain adaptation in NLP and machine learning \cite{ganin2016domain,tzeng2017adversarial}.
In contrast to all the aforementioned approaches, gradual fine-tuning requires no modification to existing models or learning objectives, so it can be applied to any system.

Domain adaptation can also be achieved using curriculum learning. \citet{zhang-etal-2019-curriculum} use a curriculum learning approach to adapt a general-domain machine translation model to a target domain while also using data whose domain is unknown. Inspired by curriculum learning \citep{10.1145/1553374.1553380}, which highlights the importance of the order of training instances, we propose a multi-stage fine-tuning strategy for domain adaptation. In this work, we order a sequence of fine-tuning datasets from least similar to the target domain to most similar.

\section{Method}
Domain adaptation via fine-tuning aims to improve performance on a target domain by using information learned from general domain data when training on data in the target domain. In other words, it is expected that training on general domain data would provide better initialization for subsequently training on target domain data than random initialization would provide \citep{erhan2010does}.

\subsection{Mixed Domain Training}
Given training data for the target domain, \indomain{}, we augment it with out-of-domain data \outofdomain{} mapped to the target task schema.\footnote{Portions of the target task schema corresponding to fields not available in the out-of-domain data could be masked in the mapped data.} The in-domain and out-of-domain data are concatenated to form a mixed domain training set. A mixed domain model is obtained by training on the mixed dataset.

\subsection{One-Stage Fine-tuning}
The distribution of the mixed domain data is different from that of the target domain data. However, regardless of how diverse the mixed domain training data is, subsequently fine-tuning the model on the target in-domain data is a direct way to encourage the model to converge to the distribution of the target domain. The model is expected to use task knowledge learned from the mixed domain data to yield improved performance over a model trained only on data from the target domain. Because the model is fine-tuned to convergence on the target domain data once, we refer to this procedure as one-stage fine-tuning.

%Our experiments (Section \ref{sec:Exp}) shows that this one-step fine-tuning significantly outperforms the one which directly trains from scratch in the tested task.

\begin{algorithm}[h] 
\begin{small}
\caption{Gradual Fine-Tuning}
\label{alg:gradualfinetuning}
\begin{algorithmic}[1]
\Require{in-domain data \indomain{}, out-of-domain data \outofdomain{0}, initial model \model{0}, out-of-domain data schedule \schedule}
\Statex
\Function{Gradual-FT}{\indomain{}, \outofdomain{0}, \model{0}, \schedule}
    \State {$t$ $\gets$ {$0$}}
    \For{$amount$ in \schedule}
        \State {$t \gets t+1$}
        \State {\outofdomain{t} $\gets$ \Call{Sample}{\outofdomain{t-1}, $amount$}}
        \State {\traindata{t} $\gets$ \indomain{} $\cup$ \outofdomain{t}}
        \State {\model{t} $\gets$ \Call{Train}{\model{t-1}, \traindata{t}}}
    \EndFor
    \State \Return {\model{t}}
\EndFunction
\end{algorithmic}
\end{small}
\end{algorithm}

\subsection{Gradual Fine-tuning}
Instead of adapting the model to the target domain by one-stage fine-tuning, we propose an iterative multi-stage approach that transitions from the initial mixed domain to the target domain. Each iteration incorporates less out-of-domain data than the preceding iteration as specified by a data schedule \schedule, bringing the data distribution closer to that of the target domain every training cycle. At each iteration, the model is trained to convergence. The number of iterations and the out-of-domain data schedule \schedule~are hyperparameters that can be tuned for a particular task. Pseudocode for gradual fine-tuning is presented in Algorithm~\ref{alg:gradualfinetuning}. In our implementation, the out-of-domain data is uniformly randomly sampled from the out-of-domain data used in the previous iteration.
%  However, the out-of-domain data may be sampled from the full set each time instead.
% Gradual fine-tuning can be divided into following steps:
% \begin{itemize}
%     \item Step 1: Obtain pretrained general-domain model.
%     \item Step 2: Keep the in-domain data unchanged and reduce the size of out-of-domain data in the combined training data.
%     \item Step 3: Fine-tune the model with the smaller combined data derived by step 2.
% \end{itemize}

% Beginning with step 1, we repeat step 2 and step 3 until there is no out-of-domain data in the combined data. 

% We will illustrate our preference of selecting these parameters in Sec. \ref{sec:Exp} for various tasks.

\section{Experiments}
\subsection{Dialogue State Tracking}
Dialogue state tracking (DST) involves estimating at each dialogue turn the probability distribution over slot-values enumerated in an ontology. For example, we may be interested in the distribution over cuisines given the dialogue history and an utterance indicating the \textit{restaurant} domain and food slot. We show that gradual fine-tuning can substantially improve slot accuracy---the accuracy of predicting each slot separately---and joint accuracy---the percentage of turns in which all slots are predicted correctly---in a given dialogue domain. 

%DST includes determination  of the full representation for the goal of the user at that point in the dialogue, which contains an intention constraint, a set of requested slots, and the user's dialogue act
%\subsubsection{Dataset}
\paragraph{Dataset}
We run experiments on the MultiWOZ v2.0 dataset~\citep{budzianowski2018large}, which is a multi-domain conversational corpus with seven domains and 35 slots. Following \citet{wu-etal-2019-transferable}, we focus on five domains: \textit{restaurant}, \textit{hotel}, \textit{attraction}, \textit{taxi}, and \textit{train}, which amounts to 2198 single-domain dialogues and 5459 multi-domain dialogues from the original dataset across all data splits. Statistics for single-domain dialogues of the five domains are presented in \autoref{tab:datasize}. Among the five domains, \textit{restaurant} and \textit{hotel} are adopted as the target domains for our experiments. We consider single-domain dialogues in the target domain as in-domain data and the rest of the (single-domain or multi-domain) dialogues excluding the target domain as out-of-domain data.
% Note that the smaller augmented data is randomly sampled from its previous larger dataset, e.g., 2k dialogues are randomly sampled from 4k dialogues.

\paragraph{Settings}
The gradual fine-tuning data schedule of out-of-domain dialogues is \schedule~$=$~4K $\rightarrow$ 2K $\rightarrow$ 0.5K $\rightarrow$ 0 (K=thousand), where the in-domain data is mixed with the out-of-domain data at each stage.\footnote{Experiments starting with 2K (2K $\rightarrow$ 0.5K $\rightarrow$ 0) augmented data examples were also conducted to further explore the effect of data size (\autoref{appendix:dst-2k}).} We use the Slot-Utterance Matching Belief Tracker (SUMBT) model  by \citet{lee-etal-2019-sumbt}, as well as their hyperparameters. The SUMBT model is composed of four parts: BERT encoders for encoding slots, values, and utterances, a slot-utterance matching network, a belief tracker, and a nonparametric discriminator. SUMBT achieves state-of-the-art performance on the MultiWOZ v2.0 dataset. More training details can be found in \autoref{appendix:training-settings}.

\begin{table}
\begin{center}

\begin{small}
  \begin{tabular}{lccccc}
  \toprule
  & Rest. & Hotel & Attract. & Taxi & Train \\
  \midrule
  \# Slots & 7 & 10 & 3 & 4 & 6 \\
  \midrule
  \# Turns & 3011 & 3472 & 577 & 1667 & 1771  \\
  \# Dialogues&&&&&\\
  \quad Train & 523 & 513 & 127 & 326 & 282\\
  \quad Dev & 50 & 56 & 11 & 57 & 30\\
  \quad Test & 61 & 65 & 12 & 52 & 33\\
  \bottomrule
 
  \end{tabular}
\end{small}
\end{center}
\caption{Data statistics for five domains from MultiWOZ~v2.0.}
\label{tab:datasize}
\end{table}
%The first two row indicates the number of slots and single in-domain utterances in each domain, and the rest of rows show the number of single-domain dialogues corresponding to each tested domain.

% except for learning rates. For each step of gradual fine-tuning, we train for either 200 epochs, or until convergence measured by using a patience threshold of 15. Our strategy of learning rates setting was determined empirically from preliminary experiments and determined by the tasks. Typically, the learning rates are the same for each step of fine-tuning, except for a lower learning rate setting for the last step. We also tried other setting strategies, such as learning rate reduction in each step, but the preliminary results show these more complex strategies do not help. In this experiment, we set the learning rate as 4e-5 for the last step of fine-tuning, and 1e-4 for the rest.

\paragraph{Baselines}
Three baselines are considered in this experiment. The first one is a model trained only on in-domain data (no data augmentation). The second is a model trained with the one-stage fine-tuning strategy (\schedule~$=$~4K $\rightarrow$ 0). The last baseline is a model trained with the same settings as \citet{lee-etal-2019-sumbt}, which has seen the full training set.\footnote{Note that use of the full training set means the model sees multi-domain dialogues which may include the target domain. The full training set contains approximately 4K more dialogues than what we use for gradual fine-tuning.}
%(indicated as \textit{No FT} in Table \ref{tab:results:original-gradual})

% To compare our approach with one-step fine-tuning method, we also tried directly fine-tuning the model with various starting points, e.g., 4k $\rightarrow$ 0.
%SUMBT extracts slot-value information by leveraging attention mechanism over user-agent utterances, and deploy a non-parametric discriminator to predict slot values. Briefly speaking, this model is mainly composed of three parts: the pre-trained BERT used for encoding slot names, slot values and utterance pairs, a multi-head attention that retrieve relevant information of  domain-slot-type from the utterances, and a RNN tracker.

\paragraph{Result and Analysis}
The main results are shown in \autoref{tab:results:original-gradual}. Compared with the model trained without data augmentation, gradual fine-tuning yields an absolute gain of 3.6\% for slot accuracy and 15.11\% for joint accuracy in the \textit{restaurant} domain and 1.7\% slot accuracy and 5.82\% joint accuracy in the \textit{hotel} domain. Moreover, gradual fine-tuning also considerably outperforms both the one-stage strategy as well as the mixed data training.

% We further record the slot accuracy of one-step fine-tuning (depicted in dash lines). Our gradual method considerably outperforms one-step strategy, e.g., 0.83 gain for the \textit{restaurant} domain and 1.06 for the \textit{hotel} domain with 4k augmented data.

% Moreover, as indicated in \autoref{tab:results:original-gradual}, our method can reach 2.11 higher scores than the state-of-the-art model \citep{lee-etal-2019-sumbt} in the \textit{restaurant} domain, and 1.01 in the \textit{hotel} domain.

\begin{figure*}
     \centering
     \begin{subfigure}[h]{0.9\columnwidth}
         \centering
         \includegraphics[width=\textwidth]{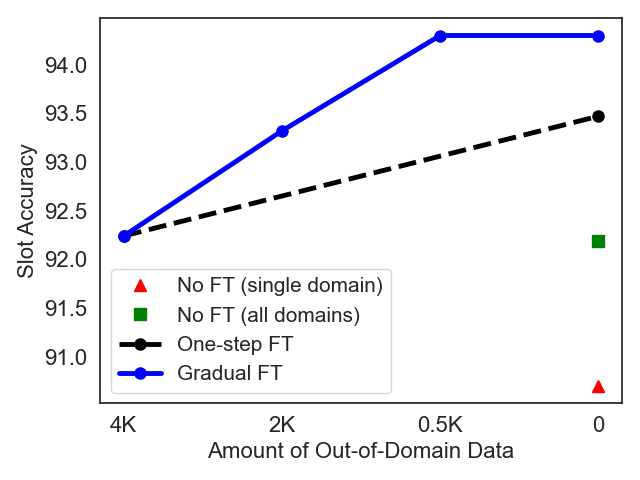}
         \caption{Restaurant}
        %  \label{fig:result-rest}
     \end{subfigure}
     %\hfill
     ~~~~~~~~~~~~~~~~~~~
     \begin{subfigure}[h]{0.9\columnwidth}
         \centering
         \includegraphics[width=\textwidth]{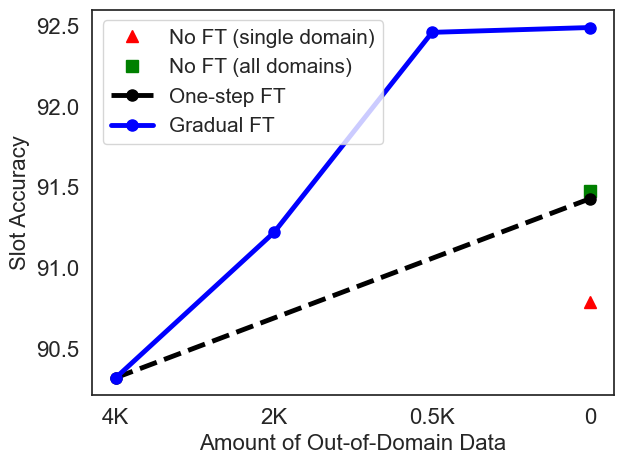}
         \caption{Hotel}
        %  \label{fig:result-hotel}
     \end{subfigure}
     \caption{Slot accuracy for the \textit{restaurant} and \textit{hotel} domains in MultiWOZ v2.0. Gradual fine-tuning yields higher performance than fine-tuning in a one-stage process. Both fine-tuning approaches perform better than not doing any fine-tuning. Results at the end of training are the rightmost points (\autoref{tab:results:original-gradual}). }
     \label{fig:results}
\end{figure*}

\autoref{fig:results} shows increasing slot accuracy at each stage of (gradual) fine-tuning, which supports our hypothesis that gradually fine-tuning the model consistently improves performance as the data distribution approaches that of the target domain.  % which also reflects our assumption that the general domain is closer to the target domain by reducing the size of out-of-domain data in each step. 

\begin{table}[h]
\begin{small}
\centering
\begin{tabular}{lcc}
% & \citet{lee-etal-2019-sumbt} & gradual fine-tuning \\
% \hline
% restaurant & 92.19 & \bf 94.30 \\
% hotel & 91.48 & \bf 92.49 \\
\toprule
& Restaurant & Hotel \\
& Slot/Joint & Slot/Joint  \\
\midrule
No FT (single domain) & 90.70/52.16 & 90.79/46.30 \\
No FT$^{*}$ (all domains) & 92.19/58.63 & 91.48/50.26 \\
One-stage FT & 93.47/61.15 & 91.43/46.30 \\
Gradual FT & \best{94.30/67.27} & \best{92.49/52.12} \\
\bottomrule

\end{tabular}
\caption{Slot and joint accuracy in the restaurant and hotel domains under various training methods. * indicates that the model uses the full training set. \textit{No FT (single domain)} is the model trained only on in-domain data and \textit{No FT (all domains)} is the training regime from \citet{lee-etal-2019-sumbt}. FT=fine-tuning.}
\label{tab:results:original-gradual}
\end{small}
\end{table}

% Complete results shown in \hl{Appendix}

\subsection{Event Extraction}
We also employ gradual fine-tuning on an event extraction task to show the general applicability of the approach. Event extraction involves predicting event triggers, event arguments, and argument roles. We perform event extraction on the ACE 2005 corpus by considering Arabic as the target domain and English as the auxiliary domain.

\paragraph{Settings}
For data processing, model building, and performance evaluation, we use the DYGIE++ framework,\footnote{\url{https://github.com/dwadden/dygiepp}} which achieved state-of-the-art results on ACE 2005 event extraction \citep{wadden-etal-2019-entity}. We replace the BERT encoder~\cite{devlin-etal-2019-bert} with XLM-R \citep{conneau-etal-2020-unsupervised} to train models on monolingual and mixed bilingual datasets. Because no standard splits were found for the Arabic portion of the ACE 2005 dataset, the train/dev/test splits for Arabic are randomly selected. \autoref{tab:EE:datasize} shows statistics of the ACE 2005 events dataset for our experiments. The gradual fine-tuning data schedule of preprocessed English documents, \schedule, is 1K $\rightarrow$ 0.5K $\rightarrow$ 0.2K $\rightarrow$ 0.\footnote{1K, 0.5K, and 0.2K preprocessed documents correspond to 85\%, 35\%, and 5\% of total events/arguments in the English training set.} We report four metrics for the evaluation: a trigger is correctly identified if its offsets find a match in the ground truth (TrigID), and it is correctly classified if their event types match (TrigC). An argument is correctly identified if its offsets and
event type find a match in the ground truth (ArgID), and it is correctly classified if their event roles match (ArgC). 
%  Results are given in \autoref{tab:results:ace2005}.

\begin{table}[h]
\begin{small}
\centering
\begin{tabular}{lcc}
\toprule
& English & Arabic\\ 
 \midrule
\# Event types & 33 & 30\\ 
\# Role types & 22 & 21\\
\# Events/Arguments & &\\
% \multirow{3}{*}{\# Events/Args} & Train & 4202/4859 & 1743/2506 \\
\quad Train & 4202/4859 & 1743/2506 \\
\quad Dev & 450/605 & 117/174 \\
\quad Test & 403/576 & 198/287 \\ 
\bottomrule
\end{tabular}
\caption{Statistics for English and Arabic ACE 2005.}
\label{tab:EE:datasize}
\end{small}
\end{table}

%\begin{small}
%  \begin{tabular}{lcccc}
%  \toprule
%  & Ar train & Ar dev & Ar test & En train \\
%  \midrule
%  \# Documents & 187 & 60 & 50 & 1025   \\
%  \# Sentences & 1765 & 542 & 481 & 16212 \\
%  \bottomrule
%  \end{tabular}
%\end{small}

\paragraph{Baselines}
The first baseline in \autoref{tab:results:ace2005} is the model from \citet{wadden-etal-2019-entity} (with an XLM-R encoder) trained only on Arabic data, the second baseline is trained on mixed data (Arabic + 1K English data) without any fine-tuning, and the third baseline uses the one-stage fine-tuning strategy.
%(indicated as  \textit{No FT (Ar)}
\paragraph{Results}
The results are given in \autoref{tab:results:ace2005}. Mixed data training and one-stage fine-tuning achieve slight improvements over no data augmentation and even hurt performance on the TrigID and ArgID metrics. Gradual fine-tuning outperforms all baselines by a substantial margin on all four metrics, especially TrigC, which achieves an absolute gain of 5.84\% compared to the state-of-the-art model trained only on the Arabic dataset. 

\begin{table}[h]
\begin{small}
\centering
\begin{tabular}{lcccc}
\toprule
& TrigID & TrigC & ArgID & ArgC \\
\midrule
% - old results
% No FT (target) & 64.77 & 57.03 & 47.76 & 42.83 \\
% No FT (mixed) & 65.44 & 59.77 & 45.45 & 43.37 \\
% One-step FT & \best{65.70} & 58.36 & 46.48 & 43.34 \\
% Gradual FT & 65.20 & \best{61.40} & \best{47.77} & \best{46.23} \\
% - new results 
No FT (Ar) & 64.77 & 57.03 & 47.76 & 42.83 \\
No FT (mixed) & 64.12 & 59.48 & 46.57 & 43.21 \\
One-stage FT & 63.61 & 59.88 & 46.79 & 43.44 \\
Gradual FT & \best{66.29} & \best{62.87} & \best{48.11} & \best{44.21} \\
\bottomrule

\end{tabular}
%\caption{\{Trigger, argument\} $\times$ \{ID, classification\} F1 on Arabic ACE 2005. \textit{FT}=fine-tuning, \textit{Ar}=Arabic, \textit{mixed}=Arabic + 1k English.}
\caption{Identification and classification F1 scores for triggers and arguments on Arabic ACE 2005. FT=fine-tuning, Ar=Arabic, mixed=Arabic + 1K English.}
\label{tab:results:ace2005}
\end{small}
\end{table}

\section{Conclusion}
We have proposed a gradual fine-tuning technique that iteratively steers the distribution of augmented training data toward that of a target domain. Gradual fine-tuning can be straightforwardly applied to an existing codebase without changing the model architecture or learning objective. Through experiments on dialogue state tracking and event extraction tasks, we have demonstrated that gradual fine-tuning outperforms standard one-stage fine-tuning for domain adaptation.

\section*{Acknowledgments}
We thank the anonymous reviewers for their valuable comments. This work was supported in part by IARPA
BETTER (\#2019-19051600005) and DARPA KAIROS (FA8750-19-2-0034). The views and conclusions contained in this work are those of the authors and should not be interpreted as necessarily representing the official policies, either expressed or implied, or endorsements of DARPA, ODNI, IARPA, or the U.S. Government. The U.S. Government is authorized to reproduce and distribute reprints for governmental purposes notwithstanding any copyright annotation therein.

%\bibliography{anthology,eacl2021}
\bibliography{eacl2021}
\bibliographystyle{acl_natbib}

\clearpage
\appendix
\section{Learning Rate Schedule}
\label{appendix:training-settings}
\subsection{Potential Model Collapse}
If we use the same constant learning rate for each fine-tuning stage, the model may collapse. Accuracy on the train and dev sets may drop to 0 or near-zero at the beginning of a stage of gradual fine-tuning. This phenomenon is often caused by using an overly large learning rate. This suggests that one must carefully schedule the learning rates for each stage of gradual fine-tuning.

\subsection{Learning Rates for MultiWOZ v2.0}
In the MultiWOZ v2.0 experiments, we use 4e-5 for the learning rate in the last stage of fine-tuning and 1e-4 (default settings from \citet{lee-etal-2019-sumbt}) for the other stages.

\subsection{Learning Rates for ACE 2005}
The network is split into 2 parameter groups: the parameters of the XLM-R encoder and all other parameters. We set the base learning rate of XLM-R to 5e-5 and to 1e-3 for the other parameters (default settings from \citet{wadden-etal-2019-entity}). We do not change the base learning rate for the first two stages of fine-tuning. At the third stage, we reduce the learning rate of XLM-R to 1e-5 and of other parameters to 4e-4. We further reduce the learning rate of XLM-R to 8e-6 and the all other learning rates to 2e-4 for the final stage of gradual fine-tuning.

\section{Additional MultiWOZ v2.0 Results}
\label{appendix:dst-2k}
\begin{figure}
     \centering
     \begin{subfigure}[h]{0.45\textwidth}
         \centering
         \includegraphics[width=\textwidth]{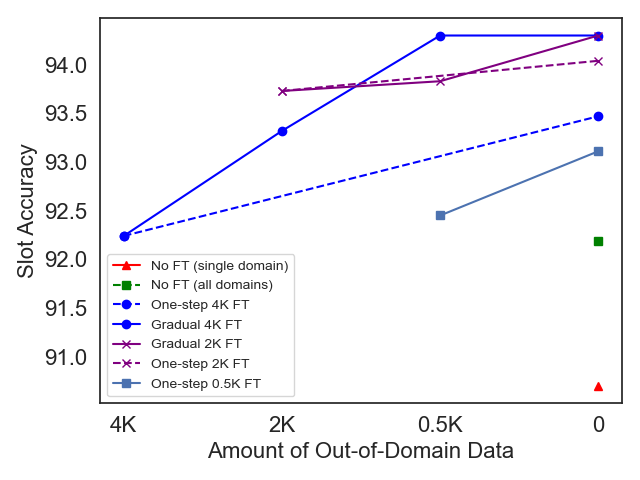}
         \caption{Restaurant}
        %  \label{fig:result-full-rest}
     \end{subfigure}
     \hfill
     \begin{subfigure}[h]{0.45\textwidth}
         \centering
         \includegraphics[width=\textwidth]{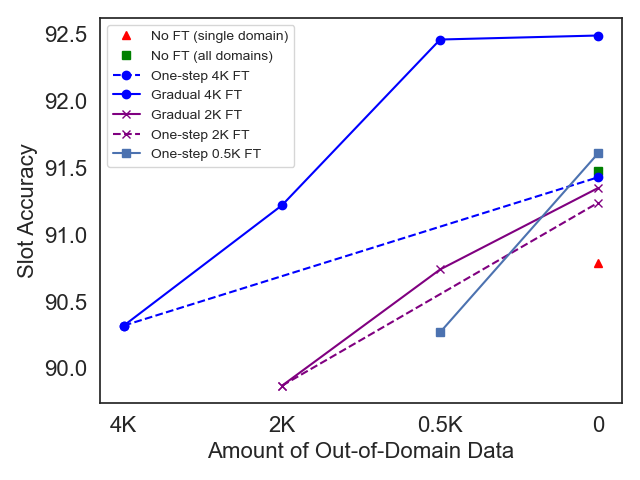}
         \caption{Hotel}
        %  \label{fig:result-full-hotel}
     \end{subfigure}
     \caption{Slot accuracy for the \textit{restaurant} and \textit{hotel} domains in MultiWOZ v2.0.}
     \label{fig:results-full}
\end{figure}

Here we present additional results of gradual fine-tuning on MultiWOZ v2.0 to augment~\autoref{tab:results:original-gradual}. \autoref{tab:results:original-gradual-2k} shows the results of one-stage fine-tuning and gradual fine-tuning using the data schedule \schedule~$=$~4K $\rightarrow$ 2K $\rightarrow$ 0.5K $\rightarrow$ 0 as well as suffixes of \schedule. \autoref{fig:results-full} shows the trend of slot accuracy for all training strategies at each stage of fine-tuning. 
% Interestingly, gradual fine-tuning with more out-of-domain data basically means higher accuracy in the final.

\begin{table}
\begin{small}
\centering
\begin{tabular}{lcc}
% & \citet{lee-etal-2019-sumbt} & gradual fine-tuning \\
% \hline
% restaurant & 92.19 & \bf 94.30 \\
% hotel & 91.48 & \bf 92.49 \\
\toprule
& Restaurant & Hotel \\
& Slot/Joint & Slot/Joint  \\
\midrule
No FT (single domain) & 90.70/52.16 & 90.79/46.30 \\
No FT$^{*}$ (all domains) & 92.19/58.63 & 91.48/50.26 \\
One-stage 0.5k FT & 93.11/64.75 & 91.61/46.30 \\
One-stage 2k FT & 94.04/62.23 & 91.24/47.35 \\
One-stage 4k FT & 93.47/61.15 & 91.43/46.30 \\
Gradual 2k FT & 94.30/66.91 & 91.35/46.30 \\
Gradual 4k FT & \best{94.30/67.27} & \best{92.49/52.12} \\
\bottomrule

\end{tabular}
\caption{Slot and joint accuracy in the \textit{restaurant} and \textit{hotel} domains under various training methods. * indicates that the model uses the full training set. \textit{FT}=fine-tuning.}
\label{tab:results:original-gradual-2k}
\end{small}
\end{table}

\end{document}